\documentclass[letterpaper, 10 pt, conference]{ieeeconf}  

\IEEEoverridecommandlockouts                              

\usepackage{graphics} 
\usepackage{epsfig} 
\usepackage{times} 
\usepackage{amsmath} 
\usepackage{amssymb}  
\usepackage{cite}
\usepackage{bm}
\usepackage{algorithm}
\usepackage[noend]{algpseudocode}
\usepackage{hyperref}
\usepackage{makecell}

\usepackage{subfigure}
\usepackage{xcolor, soul}
\usepackage{dblfloatfix}  
\usepackage{multirow}
\usepackage{nameref}
\definecolor{red}{rgb}{0.9, 0.6, 0.63}
\definecolor{cyan}{rgb}{0, 1, 1}
\DeclareMathOperator*{\argmax}{arg\,max}
\usepackage{pifont}
\newcommand{\cmark}{\ding{51}}%
\newcommand{\xmark}{\ding{55}}%

\title{\LARGE \bf
LaMOuR: Leveraging Language Models for\\Out-of-Distribution Recovery in Reinforcement Learning
}

\author{Chan Kim, Seung-Woo Seo, Seong-Woo Kim
\thanks{All authors are with Seoul National University, Seoul, South Korea,
        {\tt\small\{chan\_kim, sseo, snwoo\}@snu.ac.kr}}%
}

\begin{document}

\maketitle
\thispagestyle{empty}
\pagestyle{empty}

\begin{abstract}
Deep Reinforcement Learning (DRL) has demonstrated strong performance in robotic control but remains susceptible to out-of-distribution (OOD) states, often resulting in unreliable actions and task failure. While previous methods have focused on minimizing or preventing OOD occurrences, they largely neglect recovery once an agent encounters such states. Although the latest research has attempted to address this by guiding agents back to in-distribution states, their reliance on uncertainty estimation hinders scalability in complex environments. To overcome this limitation, we introduce Language Models for Out-of-Distribution Recovery (LaMOuR), which enables recovery learning without relying on uncertainty estimation. LaMOuR generates dense reward codes that guide the agent back to a state where it can successfully perform its original task, leveraging the capabilities of LVLMs in image description, logical reasoning, and code generation. Experimental results show that LaMOuR substantially enhances recovery efficiency across diverse locomotion tasks and even generalizes effectively to complex environments, including humanoid locomotion and mobile manipulation, where existing methods struggle. The code and supplementary materials are available at \href{https://lamour-rl.github.io/}{https://lamour-rl.github.io/}.

\end{abstract}

\section{INTRODUCTION}
Deep Reinforcement Learning (DRL) has shown remarkable success in tackling complex robotic control tasks \cite{schulman2015high,heess2017emergence,gu2017deep,akkaya2019solving}. However, similar to other deep learning-based approaches, DRL policies often struggle with out-of-distribution (OOD) states, where they may produce unreliable actions that lead to task failure. Therefore, effectively handling OOD scenarios remains a fundamental challenge in reinforcement learning (RL) research.

Several approaches have been proposed to mitigate the impact of OOD states in RL. A common strategy \cite{jiang2022uncertaintydriven, d2019exploiting, shi2024uncertainty, osband2016deep} involves exploration-based methods, which seek to minimize the occurrence of OOD states by thoroughly exploring the state space during training. Other studies \cite{kahn2017uncertainty, lutjens2019safe, tang2022prediction, henaff2018modelpredictive, kang2022lyapunov, yu2023safe, fujimoto2019off, kumar2019stabilizing, kumar2020conservative, li2022dealing} have focused on preventing the agent from entering OOD states in the first place. However, these methods primarily aim to reduce or prevent OOD occurrences rather than addressing how the agent should recover once it enters an OOD state. In real-world environments with complex and extensive state spaces, an agent will inevitably find itself in OOD situations during deployment. For example, a humanoid robot trained to walk without falling may unexpectedly slip on a wet floor and fall, entering an unintended OOD state. Therefore, it is crucial to develop methods that enable the agent to effectively recover when such situations arise.

\begin{figure}[t]
    \centering
    \includegraphics[width=\linewidth]{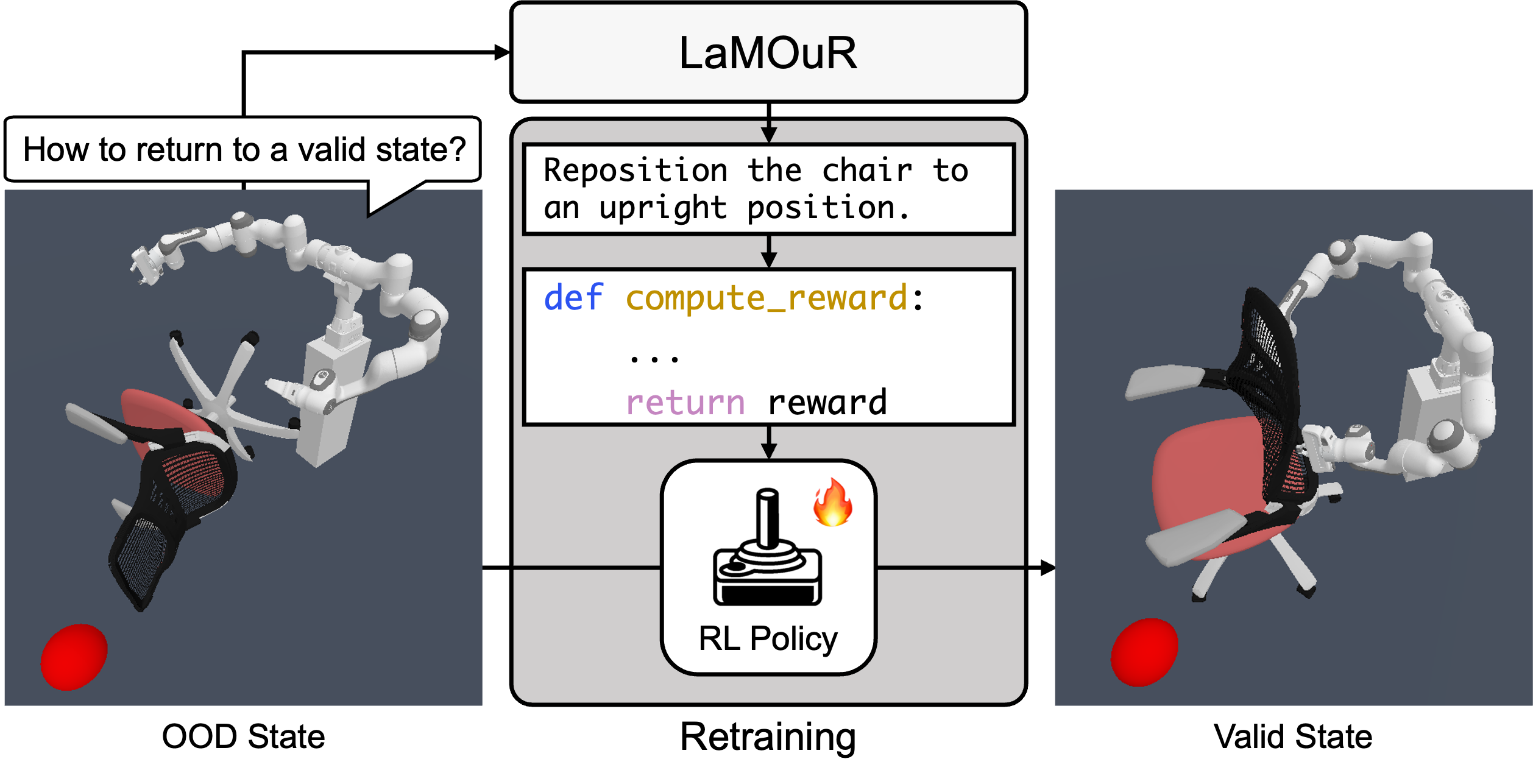}\\[-1.1ex]
    \caption{LaMOuR generates dense reward code to guide the agent’s recovery from an OOD state. Retraining with this reward enables the agent to return to states where it can successfully perform the original task.} \vspace{-1.5em}
\label{fig_concept}
\end{figure}

To address this challenge, \cite{kim2023sero} proposed a method that retrains the agent to return to an in-distribution state from OOD situations. This approach utilizes an auxiliary reward based on epistemic uncertainty, demonstrating that the agent can effectively learn to recover from OOD states. However, this method heavily depends on uncertainty estimation accuracy, making it difficult to apply in complex environments where uncertainty prediction performance degrades.

To overcome this limitation, we propose utilizing \textbf{La}nguage \textbf{M}odels for \textbf{Ou}t-of-Distribution \textbf{R}ecovery (LaMOuR) to enable agent re-learning for recovering from OOD states. Rather than relying on uncertainty estimation, LaMOuR leverages the capabilities of large vision-language models (LVLMs) to generate dense recovery reward codes that help the agent recover from OOD situations and return to a state where it can resume its original task, as shown in Fig. \ref{fig_concept}. This approach begins by utilizing an LVLM to generate a description of the agent’s OOD state based on visual inputs, such as environment-captured images. Then, using both the description of the original task and the generated OOD state description, the LVLM infers the necessary actions for the agent to return to a state where it can successfully perform its original task. Finally, based on the inferred action, the LVLM generates dense reward codes that facilitate the agent's recovery. Additionally, to prevent catastrophic forgetting of the original task during recovery learning, we introduce language model-guided policy consolidation. For clarity and simplicity, we refer to the state in which the agent can successfully perform the original task as the valid state throughout the remainder of the paper.

Contributions are summarized as follows:
\begin{itemize}
  \item We propose a novel framework that leverages the capabilities of LVLMs to retrain agents for recovering from OOD states.
  \item Experiments across diverse locomotion tasks show that our method enables more efficient recovery from OOD states compared to existing approaches.
  \item Our approach generalizes well to complex environments, such as humanoid locomotion and mobile manipulation, where prior methods often fail.
\end{itemize}

\section{RELATED WORK}
\subsection{Addressing OOD in RL}
Research on uncertainty-driven exploration \cite{jiang2022uncertaintydriven, d2019exploiting, shi2024uncertainty, osband2016deep} aims to reduce OOD states by encouraging agents to explore high-uncertainty regions, thereby maximizing state coverage. Similarly, robust adversarial RL \cite{pinto2017robust, pan2019risk, zhang2021robust} introduces adversarial noise during training to augment the agent’s encountered state space, enabling the learning of more robust policies. However, in environments with vast and complex state spaces, neither exploration nor state augmentation alone can encompass all possible states, making it practically impossible to completely eliminate OOD states.

To address this limitation, several methods have been proposed to prevent agents from entering OOD states \cite{kahn2017uncertainty, lutjens2019safe, tang2022prediction, henaff2018modelpredictive, kang2022lyapunov, yu2023safe, fujimoto2019off, kumar2019stabilizing, kumar2020conservative, li2022dealing}. Kahn \textit{et al.} \cite{kahn2017uncertainty} and L\"utjens \textit{et al.} \cite{lutjens2019safe} introduced approaches that leverage model-based RL and model predictive control (MPC) to prevent agents from visiting states with high model uncertainty. Henaff \textit{et al.} \cite{henaff2018modelpredictive} proposed a method that penalizes actions leading to high-uncertainty states in model-based RL environments, thereby restricting OOD state visits. Additionally, Kang \textit{et al.} \cite{kang2022lyapunov} introduced the Lyapunov density model to prevent distribution shifts in policy, ensuring that the agent remains within the learned distribution. In offline RL, several studies \cite{yu2023safe, fujimoto2019off, kumar2019stabilizing, kumar2020conservative, li2022dealing} have proposed methods to regularize policies, preventing them from visiting states outside the offline dataset.

However, most of these approaches focus on reducing or preventing OOD states rather than providing concrete solutions for handling OOD situations when they occur during deployment. To enable agents to learn robust behaviors in OOD scenarios, Kim \textit{et al.} \cite{kim2022unicon} proposed a method that conditions policy learning on uncertainty. While this approach demonstrated a certain level of robustness in OOD situations, its performance gradually degraded as the agent encountered states further from the training distribution. Kim \textit{et al.} \cite{kim2023sero} argued that instead of taking unreliable actions in OOD situations, agents should relearn how to return to an in-distribution state where they can resume their original tasks. To achieve this, an auxiliary reward based on uncertainty estimation was introduced, demonstrating that agents could successfully relearn recovery strategies across various locomotion tasks. However, since this auxiliary reward relies on the accuracy of uncertainty estimation, its applicability is limited in complex environments where estimating uncertainty is challenging.

\subsection{Language Models for Reward Generation}
Several approaches have been proposed to generate rewards that guide agents to learn behaviors aligned with a given task's language description. Rocamonde \textit{et al.} \cite{rocamonde2024visionlanguage} leverages a pre-trained vision-language model, CLIP \cite{radford2021learning}, to compute the cosine similarity between the task description and the agent’s visual state, using this similarity as a reward. This method has proven effective in enabling agents to learn behaviors that align with language-specified tasks.

Wang \textit{et al.} \cite{wang2024} and Zeng \textit{et al.} \cite{zeng2024learning} utilize LVLMs to replace human preference selection and train reward models through preference learning \cite{christiano2017deep}. Meanwhile, \cite{pmlr-v229-yu23a, ma2024eureka, xie2024textreward} propose generating reward codes that correspond to task descriptions by utilizing the coding capabilities of large language models (LLMs). Yu \textit{et al.} \cite{pmlr-v229-yu23a} integrates predefined function-based reward generation with well-designed MPC to generate robotic behaviors. Ma \textit{et al.} \cite{ma2024eureka} and Xie \textit{et al.} \cite{xie2024textreward} proposed methods for generating free-form dense reward functions from task descriptions, achieving performance comparable to, or even surpassing, that of human-designed rewards. Notably, \cite{ma2024eureka} shows that iteratively refining reward codes sampled by LLMs enables agents to learn pen spinning, which is challenging even with manually crafted rewards.

\section{PRELIMINARIES}

\subsection{Markov Decision Process}
A Markov Decision Process (MDP) models the sequential decision-making process of agents interacting with the environment. It is formally defined as a tuple $(\mathcal{S}, \mathcal{A}, P, r, \mu, \gamma)$, where $\mathcal{S}$ represents the state space, $\mathcal{A}$ is the action space, $P: \mathcal{S} \times \mathcal{A} \times \mathcal{S} \rightarrow [0,1]$ is the transition probability distribution $P(s_{t+1}|s_t, a_t)$, $r$ is the reward function, $\mu$ is the initial state distribution, and $\gamma$ is the discount factor. The main objective of an MDP is to determine an optimal policy $\pi^*$ that maximizes the expected cumulative reward:
\begin{equation}
\pi^* = \argmax_{\pi} \mathbb{E}_{s_0,a_0,s_1,a_1,...}\left[{\sum_{t=0}^\infty \gamma^t \cdot r(s_t, a_t)}\right],
\end{equation}
where the initial state $s_0 \sim \mu(s_0)$, the action $a_t \sim \pi(a_t|s_t)$, and the next state $s_{t+1} \sim P(s_{t+1}|s_t, a_t)$.

\begin{figure*}[t]
    \centering
    \includegraphics[width=1.0\linewidth]{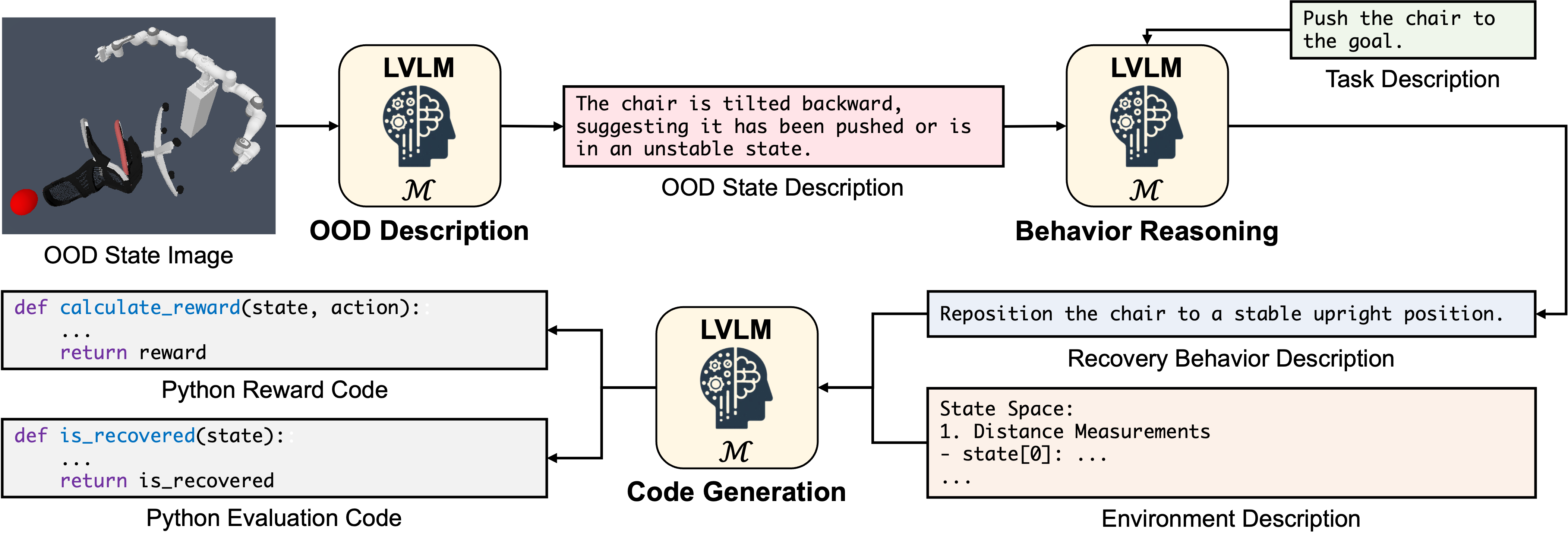}\\[-1.1ex]
    \caption{Overview of recovery reward generation in LaMOuR: First, the LVLM generates a description of the OOD state based on the given OOD state image. Next, it infers the recovery behavior needed for the agent to return to a valid state. Finally, the LVLM generates reward and evaluation codes that align with the inferred recovery behavior.} \vspace{-1.5em}
\label{fig_arch}
\end{figure*}

\subsection{Soft Actor-Critic}
Soft actor-critic (SAC) \cite{pmlr-v80-haarnoja18b} is an off-policy actor-critic method built on the maximum entropy RL framework, which seeks to maximize both the expected reward and entropy. The incorporation of maximum entropy enhances exploration and robustness, enabling SAC to address the challenges of high sample complexity and unstable convergence often faced in model-free RL. This method introduces soft value iteration, which alternates between soft policy evaluation to update the Q-function and soft policy improvement to update the policy. The parameterized Q-function $Q_\theta$ and the policy $\pi_\phi$ are updated by minimizing the following objectives:
\begin{align}
J(\theta) &= \mathbb{E}_{s_t,a_t}\left[\frac{1}{2}\left(Q_\theta(s_t,a_t)-y\right)^2\right],\label{eq_q_update}\\
J(\phi) &= \mathbb{E}_{s_t,a_t}\left[\alpha\log\pi_\phi(a_t|s_t)-Q_\theta(s_t,a_t)\right],\label{eq_pi_update}
\end{align}
with $y=r_t+\gamma\mathbb{E}_{s_{t+1}, a_{t+1}}\left[Q_\theta(s_{t}, a_{t})-\log\pi_\phi(a_{t}|s_{t}))\right]$, where $\alpha$ is entropy coefficient, $\theta$ and $\phi$ are the parameters of the Q-function and the policy, respectively.

\section{METHOD}

\subsection{Prerequisites}
To focus this study on the agent's relearning process for transitioning from an OOD state back to a valid state, we make two key assumptions. First, we assume that the agent has already learned the original task and possesses a corresponding policy, denoted as $\pi_{\phi_{org}}$. Second, we assume that the agent is already in an OOD state and is aware of it; thus, we do not address the detection of OOD situations in this study. These assumptions align with the prerequisite established in prior work \cite{kim2023sero}. The original policy can be obtained by training the agent in an environment using the reward function for the original task. We trained the policy for the original task using five different seeds and selected the one with the best performance as the original policy.

\subsection{Recovery Reward Generation}
As shown in Fig. \ref{fig_arch}, recovery reward generation consists of three phases: 1) \textit{OOD Description}: The LVLM generates a description of the agent’s OOD state based on visual inputs, such as images captured from the environment. 2) \textit{Behavior Reasoning}: Using both the textual description of the original task and the generated OOD state description, the LVLM infers the necessary actions for the agent to return to a valid state. 3) \textit{Code Generation}: The LVLM utilizes the environment description and inferred actions to generate reward code, enabling the agent to recover from the OOD state effectively. We used the same LVLM across all three modules, specifically GPT-4o \cite{gpt4o2024}.

\subsubsection{OOD Description}
To describe the agent’s state in an OOD scenario, the LVLM is given a third-person snapshot image from a fixed perspective, denoted as $I_{OOD}$, along with a prompt $p_{OOD}$. The prompt guides the model in accurately capturing the agent’s condition as depicted in the image. The generated textual description is denoted as $d_{OOD}$:
\begin{equation}
d_{OOD}=\mathcal{M}(I_{OOD},p_{OOD}).
\end{equation}

\subsubsection{Behavior Reasoning}
To infer the appropriate recovery behavior, the LVLM takes as input the original task description $d_{task}$, OOD state description $d_{OOD}$, and a prompt $p_{br}$. Based on these inputs, the model infers the recovery behavior $d_{recovery}$, which specifies how the agent should transition from the OOD state back to a valid state:
\begin{equation}
d_{recovery}=\mathcal{M}(d_{OOD},d_{task},p_{br}).
\end{equation}
We employ chain-of-thought prompting \cite{wei2022cot} to guide the LVLM in accurately inferring the appropriate recovery behavior. The model is first prompted to identify a valid state from which the agent can successfully perform the original task. Then, given the OOD state described in $d_{OOD}$, the LVLM infers the recovery behavior that enables the agent to transition from the OOD state back to a valid state.

\subsubsection{Code Generation}
Given the recovery behavior description $d_{recovery}$ and a prompt $p_{cg}$, the LVLM generates two types of code: (1) a dense reward code $c_{reward}$ that reinforces the recovery behavior and (2) an evaluation code $c_{eval}$ that determines whether the agent’s current state allows for successful task execution. To ensure that the reward generation process aligns with the environment, an environment description $d_{env}$ is provided. This allows the LVLM to generate environment-specific reward codes:
\begin{equation}
c_{reward}, c_{eval} = \mathcal{M}(d_{recovery},d_{env},p_{cg}).
\end{equation}
$d_{env}$ contains information about the agent's state and action. The prompt $p_{cg}$ provides guidance for the LVLM when generating the $c_{reward}$ and $c_{eval}$ codes. Moreover, in complex environments such as mobile manipulation, a few-shot example is included in $p_{cg}$ to assist in reward code generation. This few-shot example is not directly related to the dense reward code of the recovery behavior that the LVLM must generate, and only a single example is used.

$c_{reward}$ is designed to return the reward based on the agent's state and action as inputs. $c_{eval}$ takes the agent's state as input and evaluates whether the agent can successfully perform the original task in that state, returning one if the task can be performed and zero otherwise. Finally, the reward function used for the relearning process is defined as follows:
\begin{equation}
r_t= \begin{cases}
r^{task} & \textrm{if}\ c_{eval}(s_t) = 1\\
\lambda c_{reward}(s_{t+1},a_t) & \textrm{if}\ c_{eval}(s_t) = 0
\end{cases},
\label{eq_reward}
\end{equation}
where $r^{task}$ represents the reward provided by the environment for the original task and $\lambda$ is the scaling coefficient. The objective of this reward function is to enable the agent to learn how to return to a valid state using the recovery reward when it is in a state where task execution is not possible ($c_{eval}(s_t) = 0$). Once the agent successfully returns to a valid state ($c_{eval}(s_t) = 1$), it continues learning the original task using the task reward provided by the environment. All prompts and a few-shot example used in our method are provided in the supplementary material\footnote{Available at the project page: \href{https://lamour-rl.github.io/}{https://lamour-rl.github.io/}}.

\begin{figure}[t]
    \centering
    \subfigure[Ant]{\includegraphics[width=0.24\linewidth]{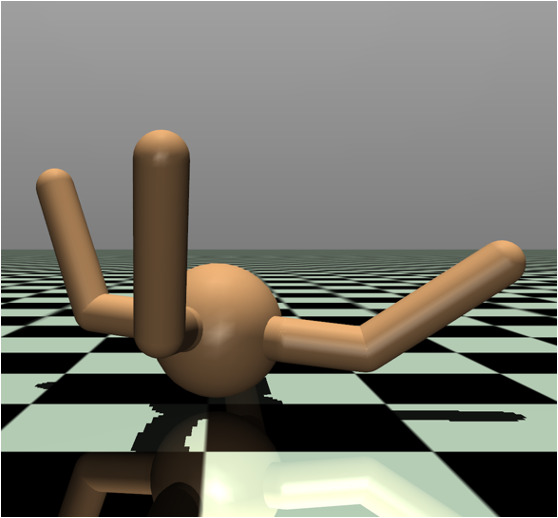}}
    \subfigure[HalfCheetah]{\includegraphics[width=0.24\linewidth]{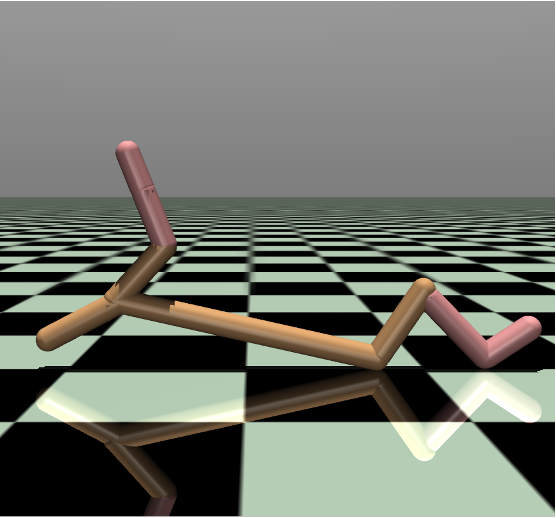}}
    \subfigure[Hopper]{\includegraphics[width=0.24\linewidth]{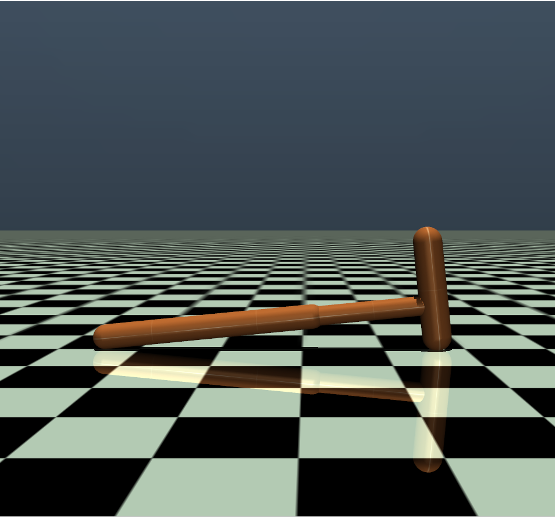}}
    \subfigure[Walker2D]{\includegraphics[width=0.24\linewidth]{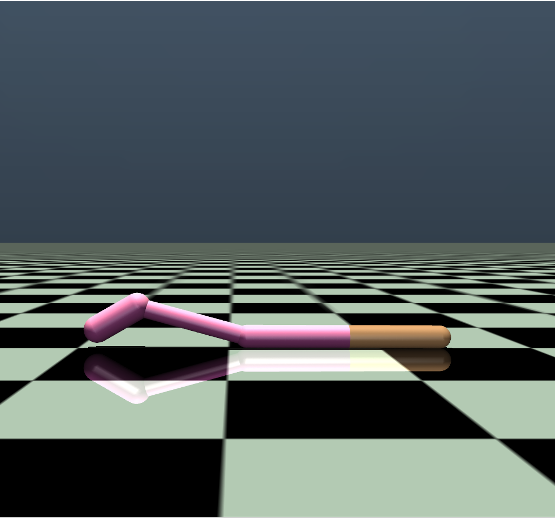}}
    \subfigure[Humanoid]{\includegraphics[width=0.475\linewidth]{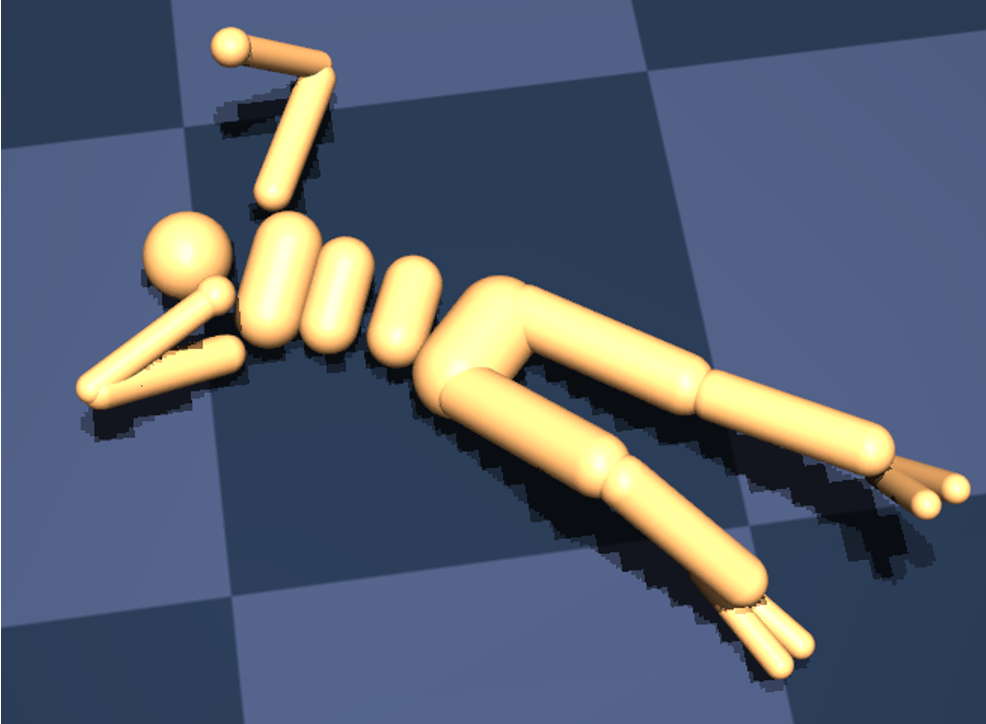}}\hspace{0.03\linewidth}
    \subfigure[PushChair]{\includegraphics[width=0.475\linewidth]{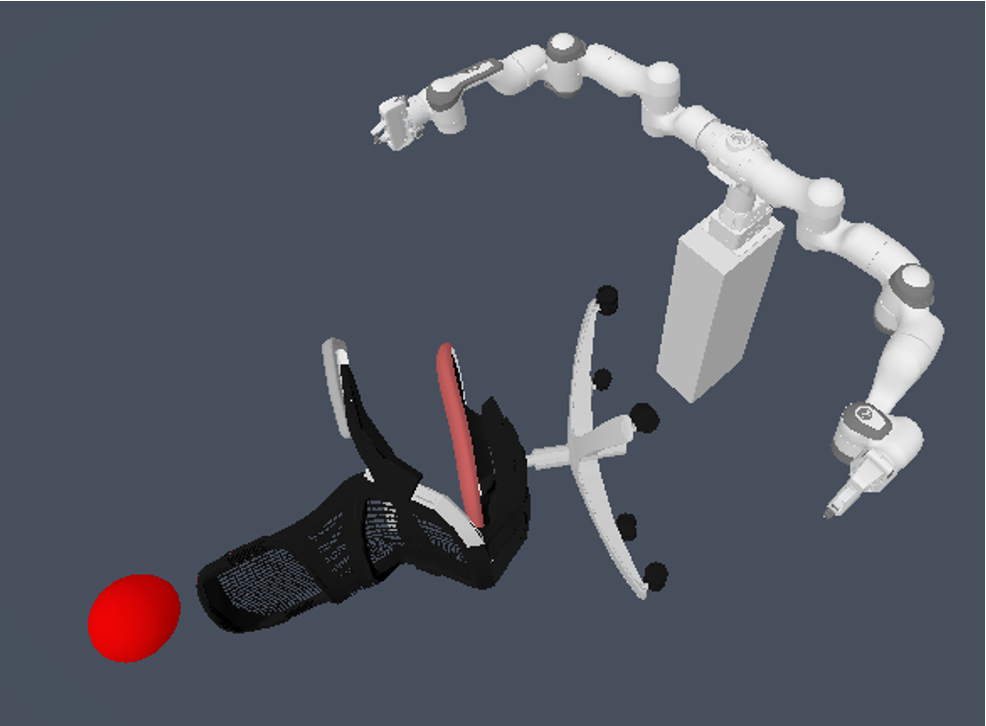}}\\[-1.1ex]
    \caption{OOD states in four modified MuJoCo and Complex environments.} \vspace{-1.2em}
    \label{fig_ood_states}
\end{figure}

\subsection{Language Model-Guided Policy Consolidation}
One way to prevent catastrophic forgetting of the original task while the agent learns to recover from an OOD state is to regularize the relearning policy $\pi_\phi$ against the original policy $\pi_{\phi_{org}}$ \cite{pmlr-v97-kaplanis19a}. However, since $\pi_{\phi_{org}}$ is likely to produce unreliable actions in OOD states, such regularization may hinder $\pi_\phi$ from effectively learning the recovery behavior. To address this issue, we propose language model-guided policy consolidation (LPC) as follows:
\begin{equation}
\mathcal{L}_{LPC} = c_{eval}(s_t) D_{KL}(\pi_\phi(a_t|s_t)||\pi_{\phi_{org}}(a_t|s_t)).
\end{equation}
Specifically, if the current state is deemed incapable of performing the original task ($c_{eval}(s_t) = 0$), policy consolidation is not applied, allowing $\pi_\phi$ to effectively learn the recovery behavior. Conversely, if the agent's current state is a valid state ($c_{eval}(s_t) = 1$), policy consolidation is applied to prevent catastrophic forgetting. Finally, the policy is trained using SAC by minimizing the following objective function, incorporating the LPC loss into Eq. (\ref{eq_pi_update}):
\begin{equation}
    J(\phi)=\mathbb{E}_{s_t,a_t}\left[\alpha \log\pi_\phi(a_t|s_t)-Q_\theta(s_t,a_t) + \mathcal{L}_{LPC}\right],
    \label{eq_lamour}
\end{equation}
while $Q_\theta$ is trained with Eq. (\ref{eq_q_update}) using Eq. (\ref{eq_reward}) as the reward function. The full algorithm is detailed in Algorithm \ref{algo_pseudo}.

\begin{algorithm}[t]
    \caption{Retraining procedure of LaMOuR.}
\begin{algorithmic}[1]
    \State Initialize $\pi_\phi$, $Q_\theta$, replay buffer $\mathcal{D}$, and load $\pi_{\phi_{org}}$.
    \State $\pi_\phi \gets \pi_{\phi_{org}}$
    \State Capture a snapshot of the OOD state, $I_{OOD}$.
    \State $d_{OOD} \gets \mathcal{M}(I_{OOD}, p_{OOD})$
    \State $d_{recovery} \gets \mathcal{M}(d_{OOD}, d_{task}, p_{br})$
    \State $c_{reward}, c_{eval} \gets \mathcal{M}(d_{recovery}, d_{env}, p_{cg})$
    \For {each episode}
        \State $s_t \sim \mu(s_0)$
        \For {each environment step}
            \State $a_t \sim \pi_\phi(a_t \mid s_t)$
            \State $s_{t+1}, r^{task}_t = \textrm{env.step}(a_t)$
            \If {$c_{eval}(s_t) = 0$}
                \State $r_t \gets \lambda c_{reward}(s_{t+1}, a_t)$
            \Else
                \State $r_t \gets r^{task}_t$
            \EndIf
            \State $\mathcal{D} \leftarrow \mathcal{D} \cup \{(s_t, a_t, r_t, s_{t+1}, c_{eval}(s_t))\}$
        \EndFor
        \For {each gradient step}
            \State $(s_t, a_t, r_t, s_{t+1}, c_{eval}(s_t)) \sim \mathcal{D}$
            \State Update $Q_\theta$ using Eq. (\ref{eq_q_update})
            \State Update $\pi_\phi$ using Eq. (\ref{eq_lamour})
        \EndFor
    \EndFor
\end{algorithmic}
\label{algo_pseudo}
\end{algorithm}

\section{EXPERIMENTS}
Our experiments are designed to answer the following key questions: 1) Does our method enhance relearning efficiency during recovery compared to other reward-based approaches? 2) Can our method generalize to recovery tasks in complex environments, such as humanoid locomotion and mobile manipulation? 3) What is the contribution of each individual component to the generation of recovery rewards?

To address the first question, we evaluated our method in four modified MuJoCo environments introduced in \cite{kim2023sero}: Ant, HalfCheetah, Hopper, and Walker2D. In these environments, agents are retrained in OOD states, depicted in Fig. \ref{fig_ood_states}(a)-(d), which the original policy never encounters. For the second question, we tested our method in the DeepMind Control Suite Humanoid environment \cite{tassa2018dmc} and the ManiSkill2 PushChair environment \cite{gu2023maniskill}, which represent complex locomotion and mobile manipulation tasks, respectively. Finally, to address the third question, we conducted an ablation study by removing individual components of the proposed method.

\begin{figure*}[t]
    \centering
    \includegraphics[width=\linewidth]{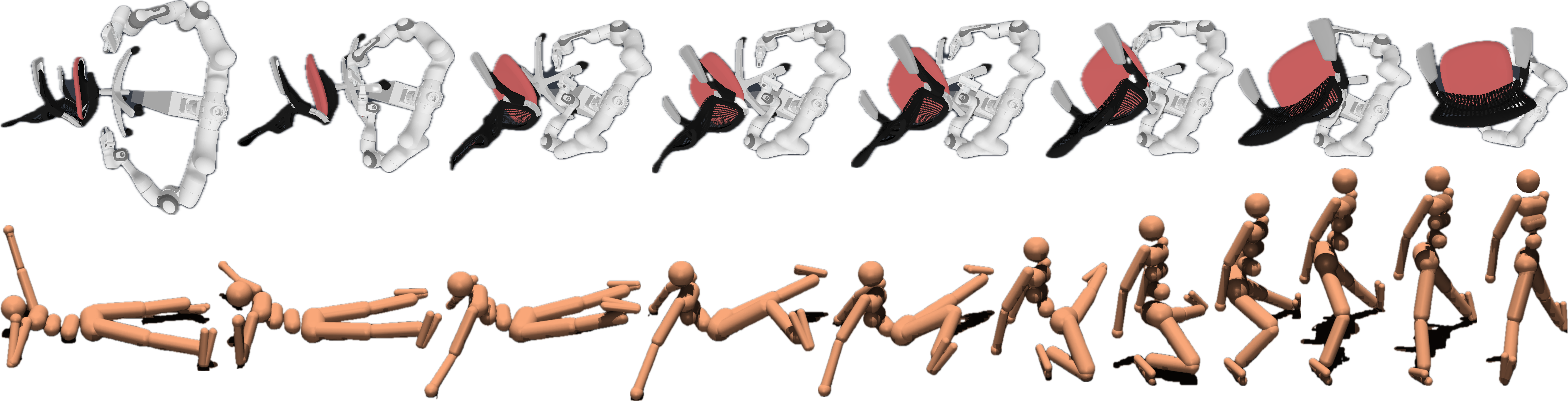} \\[-1.3ex]
    \caption{Qualitative results of the retrained agent using LaMOuR in the PushChair environment (top) and the Humanoid environment (bottom).}
    \label{fig_complex_qual}
\end{figure*}

\subsection{Evaluation in Four MuJoCo Environments}
\begin{figure*}[t]
    \centering
    \includegraphics[width=\linewidth]{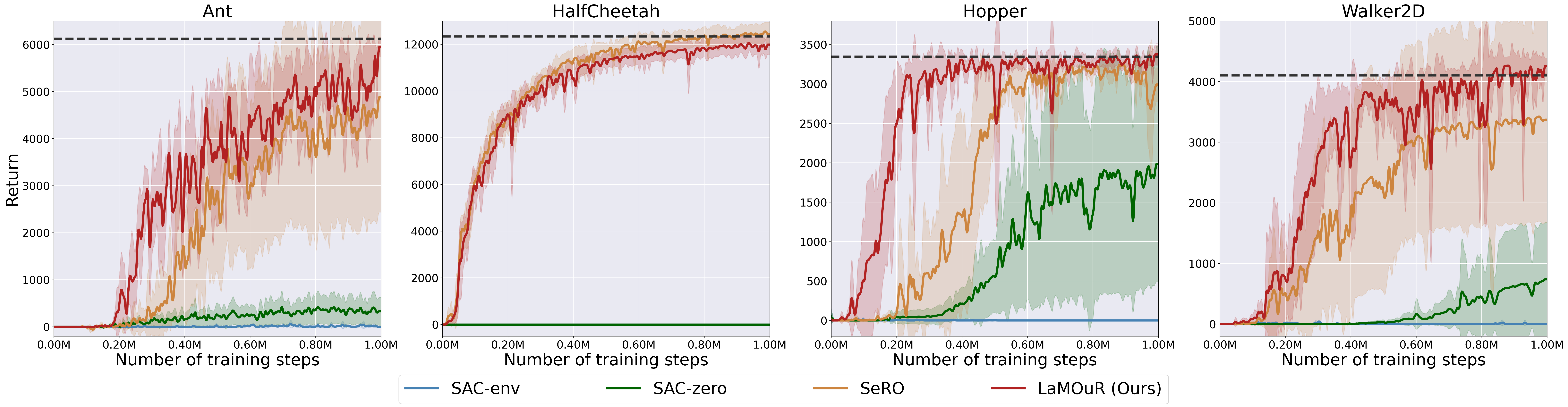} \\[-1.3ex]
    \caption{Learning curves for the four MuJoCo environments during retraining, evaluated over five episodes every 5000 training steps. Darker lines and shaded areas denote the mean returns and standard deviations across five random seeds, respectively. The black dashed line represents the average return of the original policy on the original task from a valid state. It is calculated as the average return of the original policy over 100 episodes.}
    \label{fig_mujoco_result}
\end{figure*}
We evaluated our method against three baselines: SeRO \cite{kim2023sero}, which utilizes uncertainty-based rewards in OOD states and incorporates uncertainty-aware policy consolidation; SAC-env, which uses the original task reward $r^{task}$ in OOD states; and SAC-zero, which assigns zero reward in OOD states. The learning curves during retraining are presented in Fig. \ref{fig_mujoco_result}. To provide a more intuitive representation of whether the agent successfully returns to a valid state, we plot the learning curves with the reward set to zero in OOD states.

As shown in the figure, SAC-env failed to return to a valid state, while SAC-zero struggled to recover its original performance (black dashed line), consistent with the findings in \cite{kim2023sero}. Compared to SeRO, our method demonstrates higher sample efficiency in returning to a valid state and restoring the original performance across all environments, except for HalfCheetah, where our method and SeRO exhibit similar performance. These results highlight that the dense reward signals generated by our approach offer more precise guidance for recovery, leading to improved relearning efficiency compared to uncertainty-based rewards.

\begin{figure}[t]
    \centering
    \includegraphics[width=\linewidth]{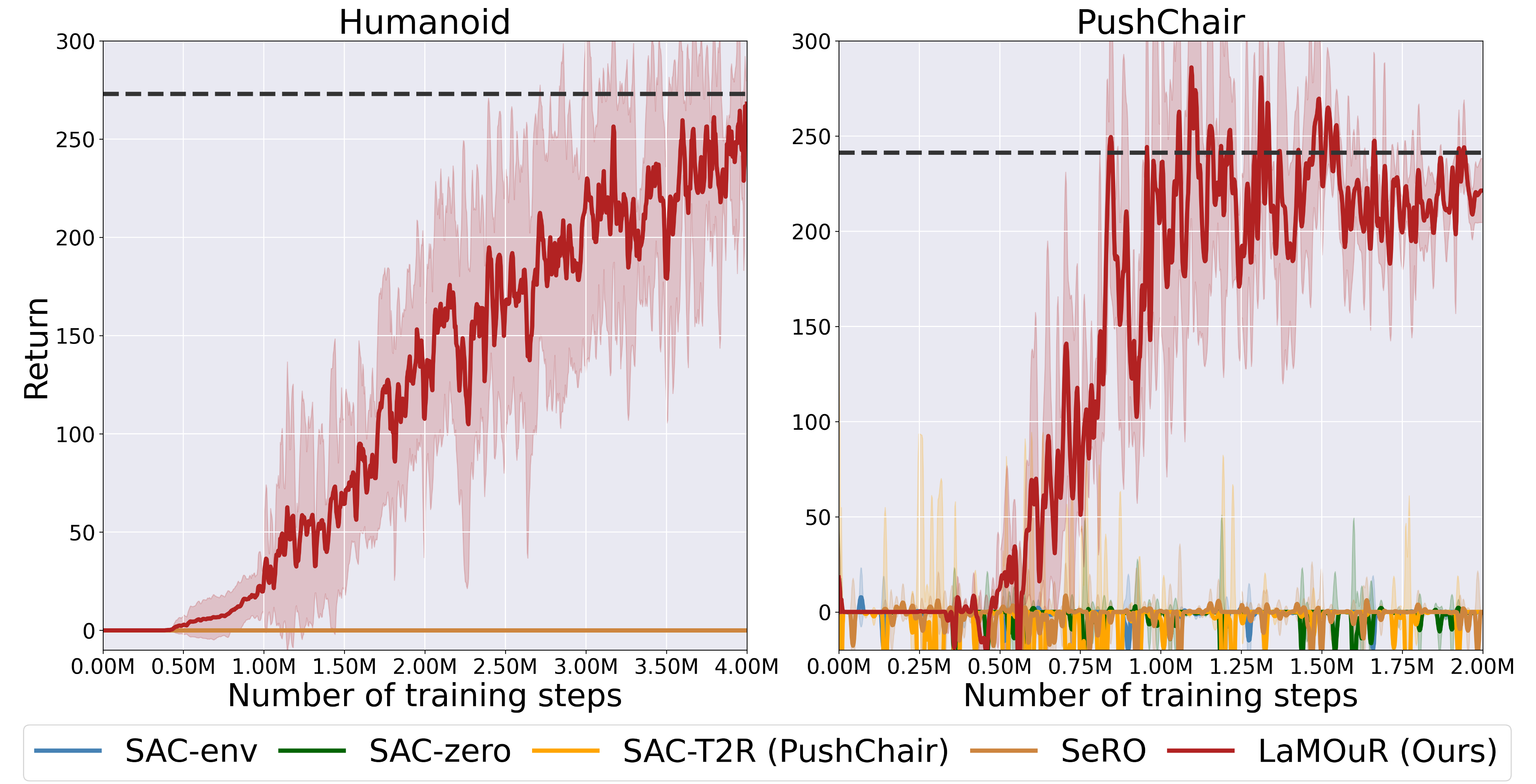} \\[-1.3ex]
    \caption{Learning curves for the two complex environments during retraining, evaluated over five episodes every 5000 training steps.}
    \label{fig_complex_result}
\end{figure}

\subsection{Evaluation in Complex Environments}
In this experiment, we compare our method against the same baselines used in the previous subsection, in both the Humanoid and PushChair environments. Additionally, for the PushChair environment, we include a SAC agent trained using rewards generated by Text2Reward \cite{xie2024textreward} (SAC-T2R) as an additional baseline. Text2Reward utilizes an LLM to generate reward functions from textual descriptions of the original task, guided by a few-shot example to produce the reward code. For this comparison, we use the generated reward function provided in the official Text2Reward repository for the PushChair environment. We do not include SAC-T2R in the Humanoid experiment due to the lack of publicly available prompts for this environment.

In the Humanoid environment, the original task is to run forward without falling. We retrained both our method and baselines from an OOD state where the humanoid is lying on the floor (Fig. \ref{fig_ood_states}(e)). In the PushChair environment, the task involves using a dual-arm Panda robot to push a chair to a target position without tipping it over. All methods were retrained from an OOD state where the chair had already fallen (Fig. \ref{fig_ood_states}(f)). These states are considered OOD because the original policies never encounter them during training, as episodes terminate when such states are reached. A few-shot example was used to generate the recovery reward for the PushChair environment. Additional environmental details are provided in the supplementary material.

\begin{table}[t]
    \centering
    \caption{Success rate in PushChair environment.}\vspace{-1em}
    \label{table_pushchair_sr}
    \resizebox{0.8\linewidth}{!}{
    \begin{tabular}{c|c c}
         & \makecell{Original $\pi_{\phi_{org}}$} & \makecell{Retrained $\pi_\phi$} \\
        \hline
        Success Rate (\%)  & 82 &  79.8\\
        \hline
    \end{tabular}
    }
\end{table}

Fig. \ref{fig_complex_result} presents the learning curves, computed using the same methodology as in the previous subsection. As shown, both SAC-env and SAC-zero fail to learn recovery behaviors. Notably, unlike in the four MuJoCo environments, SeRO fails to learn recovery in either environment. This is attributed to the degraded accuracy of uncertainty estimation in environments with more complex state and action spaces, resulting in auxiliary rewards that fail to effectively guide the agent back to valid states. In the PushChair environment, SAC-T2R also failed to learn recovery behaviors, as Text2Reward does not account for OOD states where the agent has fallen, nor does it reason about how to recover from such states. In contrast, our method successfully retrained the agent to return to a valid state and restore its original performance. This success is attributed to the dense reward code generated by our method, which provides explicit guidance on how to reach the valid state. We also evaluated the success rate of the agent retrained using LaMOuR in the PushChair environment, where success is defined as moving the chair to the goal position while keeping it upright. We assessed 100 episodes starting from OOD states across five seeds and computed the success rate. As shown in Table \ref{table_pushchair_sr}, the retrained agent achieved a success rate from OOD states comparable to that of the original policy starting from valid states.

Fig. \ref{fig_complex_qual} presents qualitative results of the retrained agent using LaMOuR in the Humanoid and PushChair environments. In the Humanoid environment, the agent first pushes against the floor with its hands to lift its upper body, then bends one knee and pushes off the ground to stand up. In the PushChair environment, the dual-arm Panda robot positions its arms near the chair base, grasps the spokes, presses down, and rotates the chair to lift it upright.

These results demonstrate that our approach effectively generalizes to complex environments, successfully training the agent to recover from OOD states and return to a valid state, where other reward-based methods fail.

\begin{figure}[t]
    \centering
    \includegraphics[width=\linewidth]{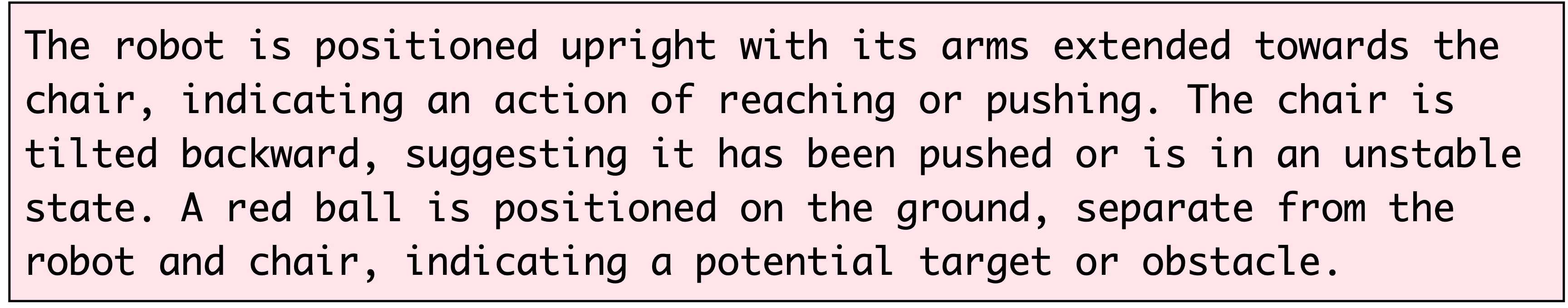}\\[-1.3ex]
    \caption{\textit{OOD Description} result for the PushChair environment.}
\label{fig_ood_desc_pushchair}
\end{figure}

\begin{figure}[t]
    \centering
    \includegraphics[width=\linewidth]{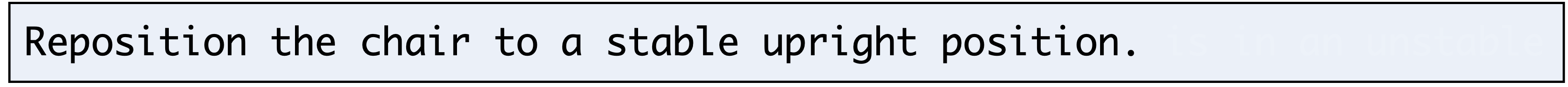}\\[-1.3ex]
    \caption{\textit{Behavior Reasoning} result for the PushChair environment.}
\label{fig_beh_rea_pushchair}
\end{figure}

\begin{figure}[t]
    \centering
    \subfigure[Evaluation code generated for the PushChair environment.]{\includegraphics[width=\linewidth]{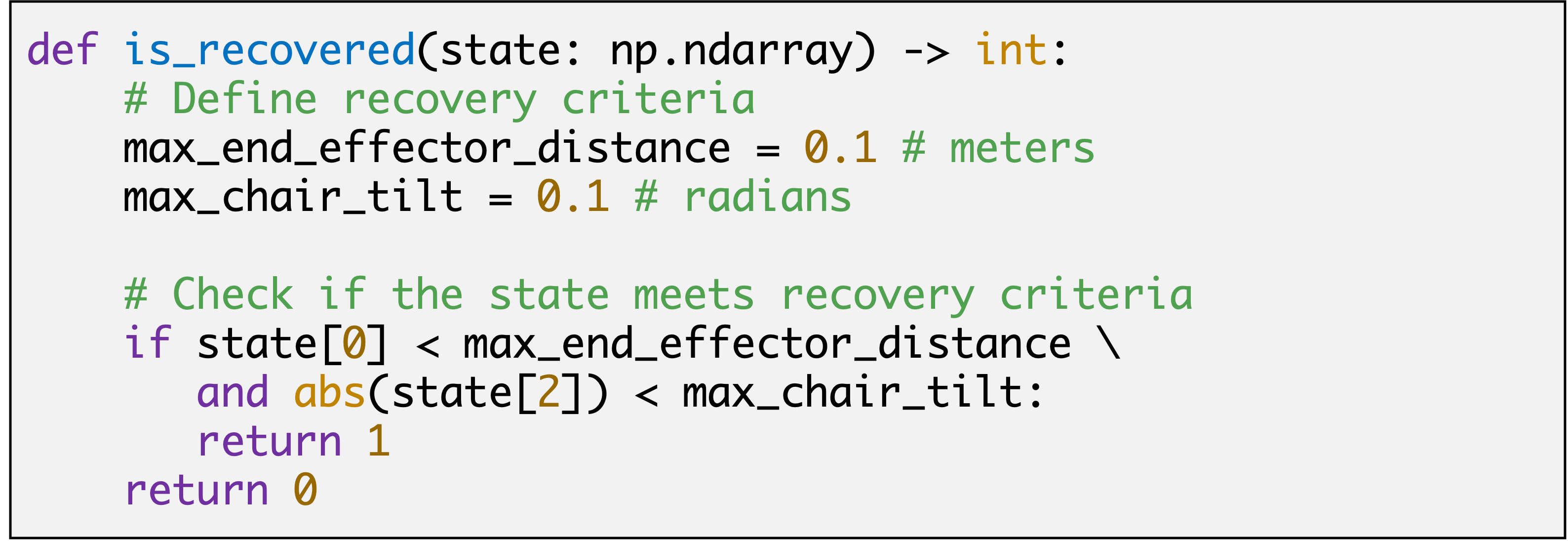}}
    \subfigure[Reward code generated for the PushChair environment.]{\includegraphics[width=\linewidth]{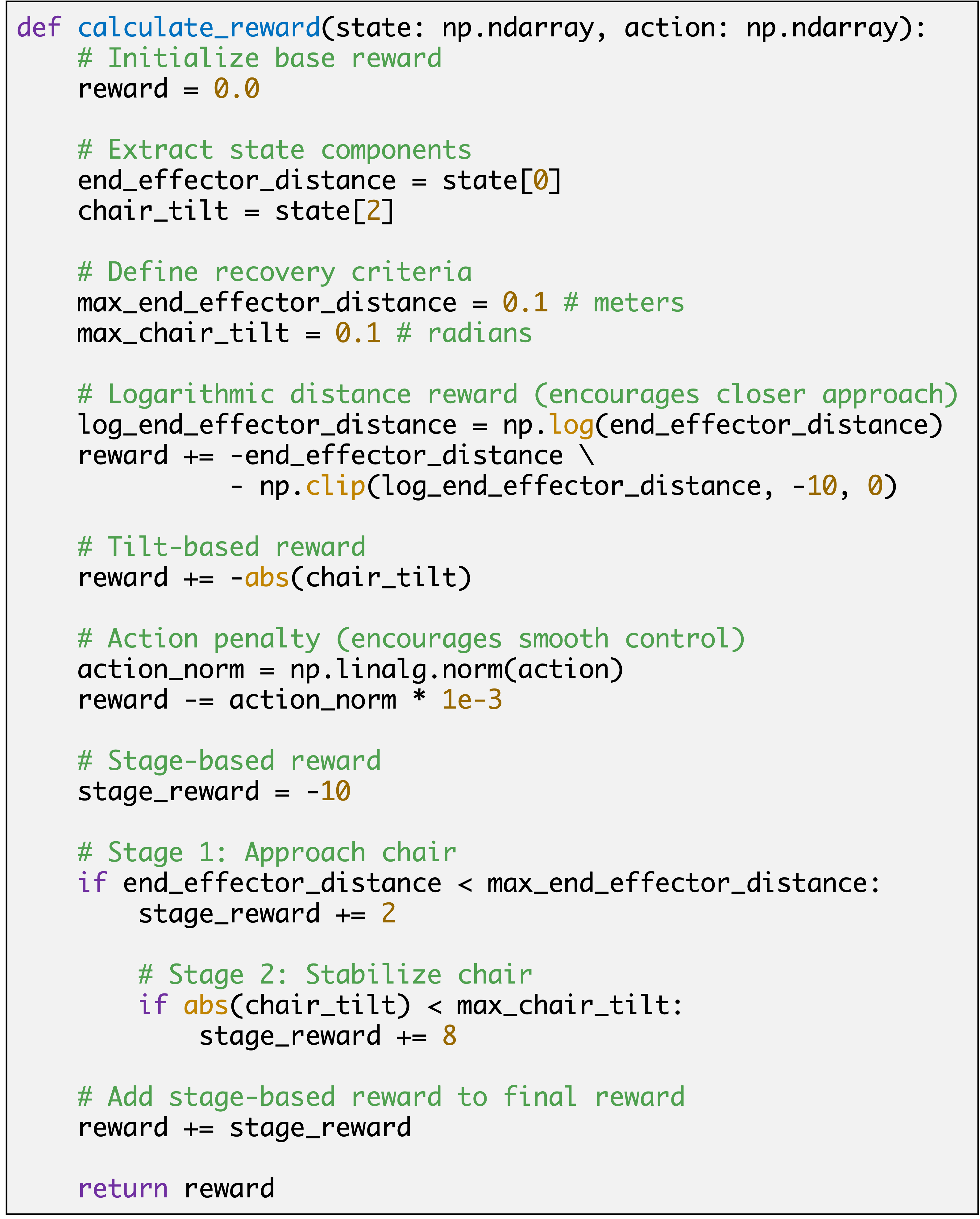}} \\[-1.3ex]
    \caption{\textit{Code Generation} results for the PushChair environment.} \vspace{-1.5em}
    \label{fig_code_gen}
\end{figure}

\begin{figure*}[t]
    \centering
    \subfigure[A1]{\includegraphics[width=0.48\linewidth]{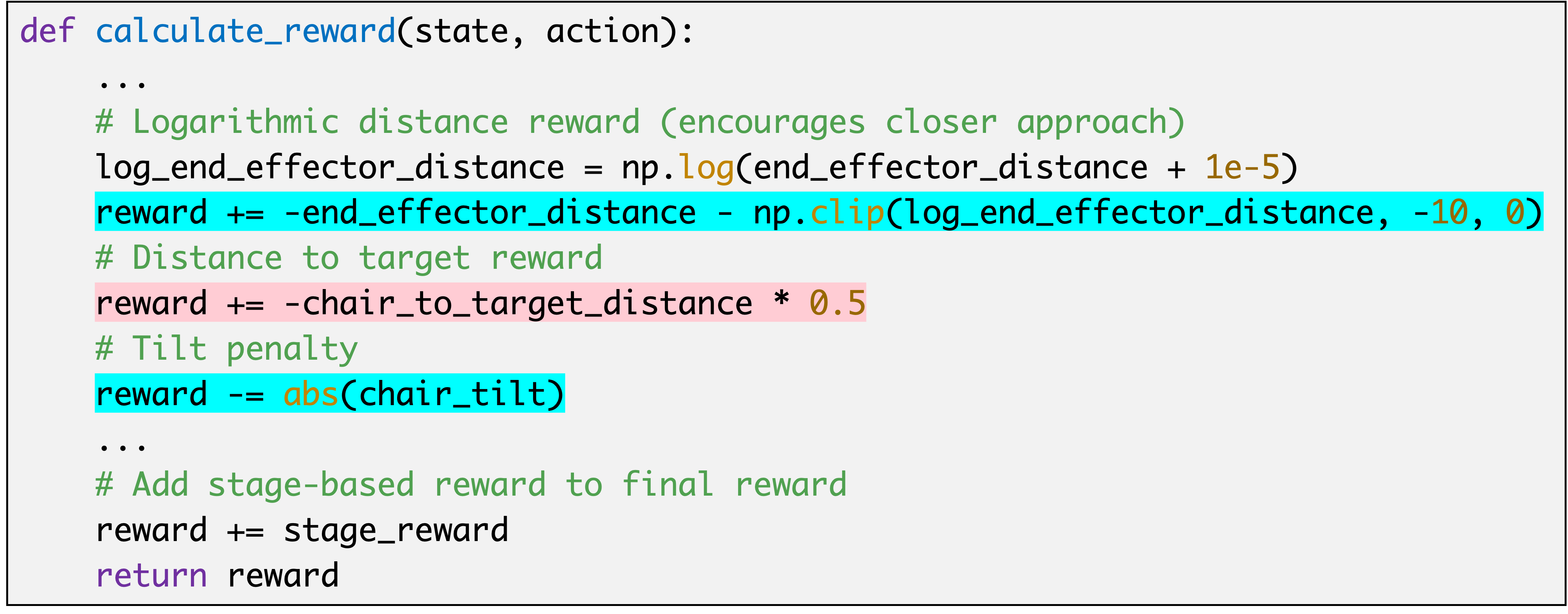}}\hspace{0.01\linewidth}
    \subfigure[A2]{\includegraphics[width=0.48\linewidth]{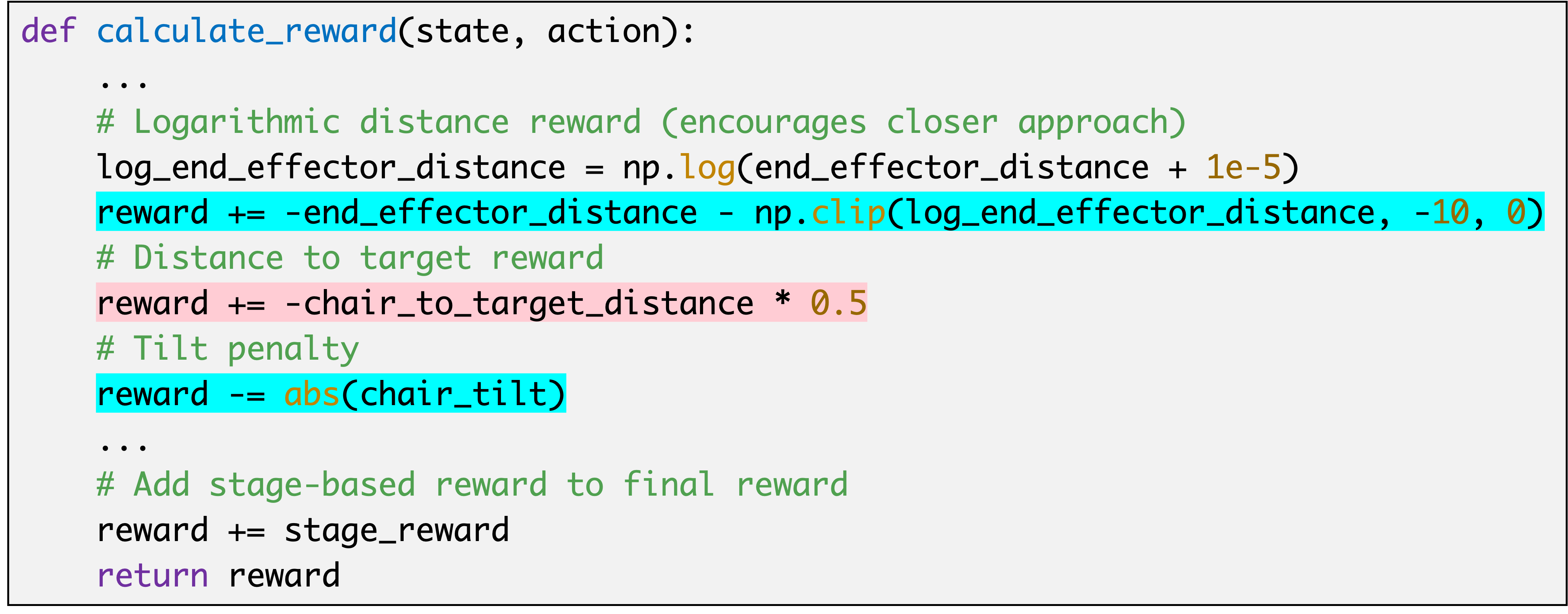}}\\[-1.1ex]
    \subfigure[A3]{\includegraphics[width=0.48\linewidth]{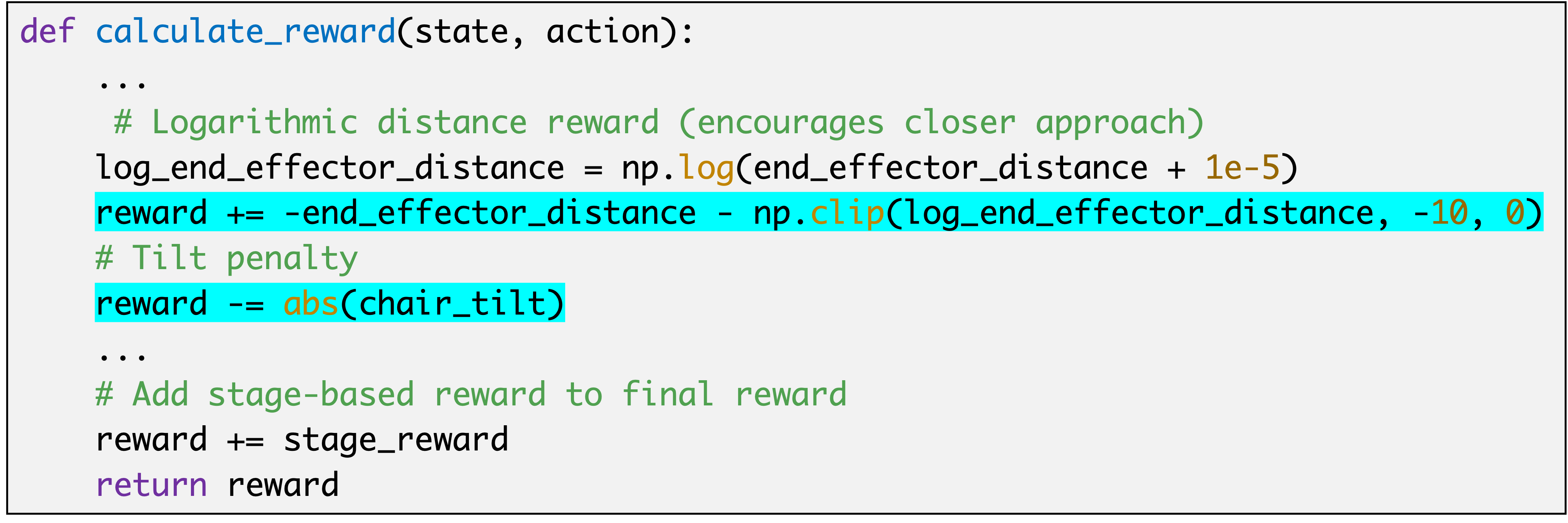}}\hspace{0.01\linewidth}
    \subfigure[A4]{\includegraphics[width=0.48\linewidth]{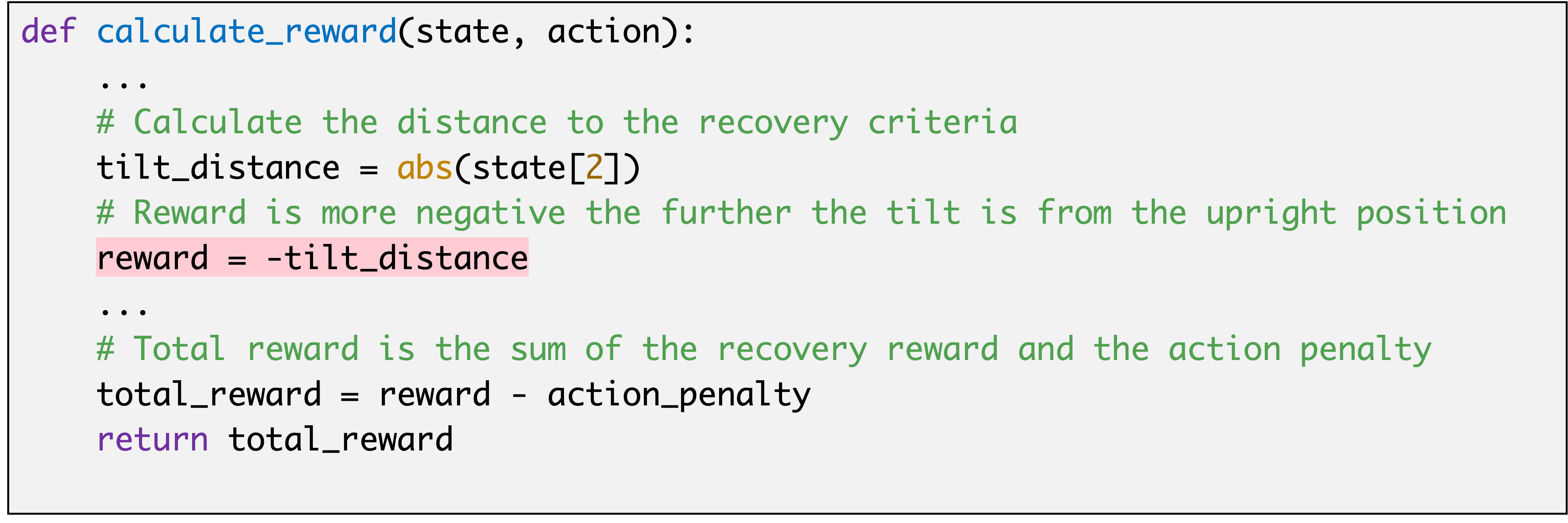}}\\[-1.1ex]
    \caption{Qualitative results of the generated code from the ablation study. Note that some parts of the code have been omitted for simplicity.}
    \label{fig_ablation}
\end{figure*}

\subsection{Qualitative Analysis of Recovery Reward Generation}
We present a qualitative example of recovery reward generation in the ManiSkill2 PushChair environment. As shown in the OOD state depicted in Fig. \ref{fig_ood_states}(f), the desired recovery behavior of the agent is to approach the chair and reposition it to an upright position. 

As shown in Fig. \ref{fig_ood_desc_pushchair}, the \textit{OOD Description} module generates a description that accurately represents the OOD state depicted in Fig. \ref{fig_ood_states}(f), illustrating the state of the robot and the chair. Fig. \ref{fig_beh_rea_pushchair} illustrates the \textit{Behavior Reasoning} result, which infers the appropriate recovery behavior of repositioning the chair, a necessary step for transitioning from the OOD state back to a valid state. Finally, \textit{Code Generation} produces both the evaluation code and the reward code, as shown in Fig. \ref{fig_code_gen}(a) and Fig. \ref{fig_code_gen}(b), respectively. The results show that \textit{Code Generation} effectively produces evaluation code based on two key factors: the distance between the dual-arm Panda robot's end-effector and the chair, and the chair’s tilt angle. It also successfully generates a reward function that increases as the agent approaches and repositions the chair. Specifically, the reward grows as the distance between the end-effector and the chair decreases, and as the chair’s tilt angle is reduced—indicating the chair is returning to an upright position. Additionally, the generated reward code encourages a staged recovery strategy by first incentivizing the robot to approach the chair, and then providing further rewards as the chair is brought back to an upright posture.

\begin{table}[ht]
    \centering
    \caption{Ablation study index}\vspace{-1em}
    \label{table_ablation}
    \resizebox{0.9\linewidth}{!}{
    \begin{tabular}{c|c c  c c}
        \hline
        Index & \makecell{\textit{OOD}\\\textit{Description}} & \makecell{\textit{Behavior}\\\textit{Reasoning}} & \makecell{\textit{Code}\\\textit{Generation}} & \makecell{Few-Shot\\Example} \\
        \hline
        A1  & \xmark  & \xmark  & \cmark & \cmark \\ \hline
        A2  & \cmark  & \xmark  & \cmark  & \cmark \\ \hline
        A3  & \xmark  & \cmark  & \cmark  & \cmark \\ \hline
        A4  & \cmark  & \cmark  & \cmark & \xmark  \\ 
        \hline
    \end{tabular}
    }
\end{table}

\subsection{Ablation Study}
We conducted an ablation study in the ManiSkill2 PushChair environment to evaluate the impact of individual components in the recovery reward generation process. Table \ref{table_ablation} outlines the various ablation settings. In A1, the OOD state image is directly provided to the \textit{Code Generation} module, whereas in A2, the generated OOD state description is supplied instead. In A3, the OOD state image is directly given to the \textit{Behavior Reasoning} module. In A4, the few-shot example is excluded during \textit{Code Generation}. The prompts for each ablation were adapted to reflect the modified input.

We found that A3, which directly provides the OOD state image to the \textit{Behavior Reasoning} module instead of using the output from an \textit{OOD Description}, also generates effective reward code (Fig. \ref{fig_ablation}(c)), comparable to the result in Fig. \ref{fig_code_gen}(b). This suggests that the \textit{Behavior Reasoning} module can effectively interpret OOD states presented either as images or as language descriptions, highlighting the strong multi-modal input processing capability of the LVLM.

However, in A1 and A2, which do not incorporate the \textit{Behavior Reasoning} module, the generated reward code does not fully align with the desired recovery behavior. As highlighted in \textcolor{red}{red} in Fig. \ref{fig_ablation}(a) and (b), the generated reward codes include terms that encourage the agent to move the chair to the target, which corresponds to the original task. While these codes include reward terms for the end-effector approaching the chair and repositioning the chair, as highlighted in \textcolor{cyan}{cyan}, the inclusion of unrelated reward terms can lead to a local minimum where the agent pushes the chair to the goal position without properly repositioning it. These results suggest that the \textit{Behavior Reasoning} module is crucial for generating the correct reward code that aligns with the desired recovery behavior.

In A4 (Fig. \ref{fig_ablation}(d)), where the few-shot example is not used for generating the reward code, the focus is solely on the tilt angle of the chair (highlighted in \textcolor{red}{red}), without accounting for how the end effector should approach the chair. This occurs because the generated recovery behavior only specifies repositioning the chair without providing guidance on reaching it, as shown in Fig. \ref{fig_beh_rea_pushchair}. While the generated reward code aligns with the intended recovery behavior, it presents a greater learning challenge for the agent. The agent can begin learning how to approach the chair only after randomly reaching it through exploration and receiving a reward for manipulating it. This result underscores the importance of a few-shot example in guiding the LLM to generate a step-by-step staged reward, particularly in environments where recovery behavior must be structured as a sequential process.

\section{CONCLUSION}
In this study, we focused on retraining agents in OOD states to learn behaviors that guide them back to states where they can successfully perform the original task. To this end, we proposed a novel approach that leverages language models to analyze OOD states, reason about recovery behaviors, and generate corresponding dense reward code. Our experiments demonstrated that the proposed method significantly improves retraining efficiency in OOD states across diverse locomotion environments. Additionally, we showed that our approach generalizes well to complex scenarios, such as humanoid locomotion and mobile manipulation, where existing methods often fall short. However, our current work addresses only the retraining aspect and does not cover the automatic detection of OOD situations during original policy execution. A promising direction for future work is to integrate an anomaly detection algorithm into our framework, enabling the agent to automatically trigger retraining when it encounters OOD states during operation.




\section*{ACKNOWLEDGMENT}
This research was supported by the Challengeable Future Defense Technology Research and Development Program through the Agency For Defense Development (ADD) funded by the Defense Acquisition Program Administration (DAPA) in 2024 (No.915027201).


\bibliographystyle{IEEEtran}
\bibliography{IEEEabrv, ref}

\clearpage
\section*{Supplementary Material}
\label{appen}

\subsection{Implementation Details}
In this subsection, we describe the implementation details of our method and baselines. For all environments, the policies and Q-functions of all methods are implemented as two-layer multi-layer perceptrons (MLPs) with 256 hidden units per layer. Each layer employs ReLU activation. All methods are optimized using the Adam optimizer \cite{kingma2015adam}. The hyperparameters used for the experiments are listed in Table \ref{table_imp_detail}. We employed the automated entropy adjustment proposed in \cite{haarnoja2018soft} for the entropy coefficient $\alpha$. Note that the same set of hyperparameters is applied across all methods.

\renewcommand{\arraystretch}{1.4}
\begin{table}[t]
    \centering
    \caption{Implementation Details}\vspace{-1em}
    \label{table_imp_detail}
    \resizebox{0.9\linewidth}{!}{
    \begin{tabular}{c c}
        \hline
        Hyperparameter & Value \\
        \hline
        Discount Factor $\gamma$ & 0.99 \ (\textrm{0.95 for PushChair}) \\
        Adam $\beta_1$ & 0.9 \\
        Adam $\beta_2$ & 0.99 \\
        Policy learning rate $\lambda_\pi$ & 0.0003 \\
        Q-function learning rate $\lambda_Q$ & 0.0003 \\
        Mini-batch size & 256 \ (\textrm{1024 for PushChair})\\
        Replay buffer size & 1M \\
        \makecell[c]{Num. of retraining steps \\ (Ant, HalfCheetah, Hopper, Walker2D)} & 1M \\
        Num. of retraining steps (Humanoid) & 4M \\
        Num. of retraining steps (PushChair) & 2M \\
        Num. of evaluation episodes & 5 \\
        Evaluation interval & 5000 \\
        Dropout rate & 0.1 \\
        Temperature of LVLMs (LaMOuR) & 0.0 \\
        \hline
    \end{tabular}
    }
\end{table}

\subsection{Environmental Details}
\subsubsection{DeepMind Control Suite Humanoid Environment}
The original task of the Humanoid environment is to control a 3D bipedal robot, to walk forward as quickly as possible without falling over. During learning the original task, the episode is terminated when the agent falls over. This termination condition is implemented by checking whether the height of the torso, $h_{torso}$, is within a healthy range. 

In our experiment, we retrain the policy in a state where the agent is lying on the floor. This state is an OOD state that the agent has never encountered while learning the original task, as the state lies outside the termination condition. The expected recovery behavior of the Humanoid in this OOD state is to stand up and return to an upright position. 

\subsubsection{ManiSkill2 PushChair Environment}
The original task of the PushChair environment is to control the dual-arm Panda robot to push the chair to the goal position. We made a slight modification to the environment, terminating the episode when the chair begins to fall over. This termination condition is implemented by checking whether the tilt angle of the chair, $\theta_{chair}$, exceeds a certain threshold. 

Accordingly, we modified the reward function of the original environment. The original reward, designed to be negative to encourage faster success, leads to agent suicide when the agent can terminate the episode by tipping the chair over, resulting in a higher cumulative reward compared to completing the task properly. Therefore, we modified the stage reward to be positive and introduced a large penalty to discourage the agent from tipping the chair over. However, making the stage reward positive no longer incentivizes fast success. Since reward design in the original environment is beyond the scope of this research, we leave it as future work. 

Moreover, we modified the state information provided to the generated reward code $c_{reward}$ because the original state from the environment lacks sufficient information for accurate reward calculation. For instance, The original state only includes the chair’s position, orientation, and velocity as its state information. However, to evaluate whether the dual arm successfully grabs the chair, it is more relevant to use the distance between the dual arm and the nearest point on the chair, rather than the distance to the chair’s center. To address this, we augment the state only when it is passed to the generated code $c_{reward}$, while the agent’s policy continues to use the original state.

\begin{figure*}[t]
    \centering
    \includegraphics[width=0.8\linewidth]{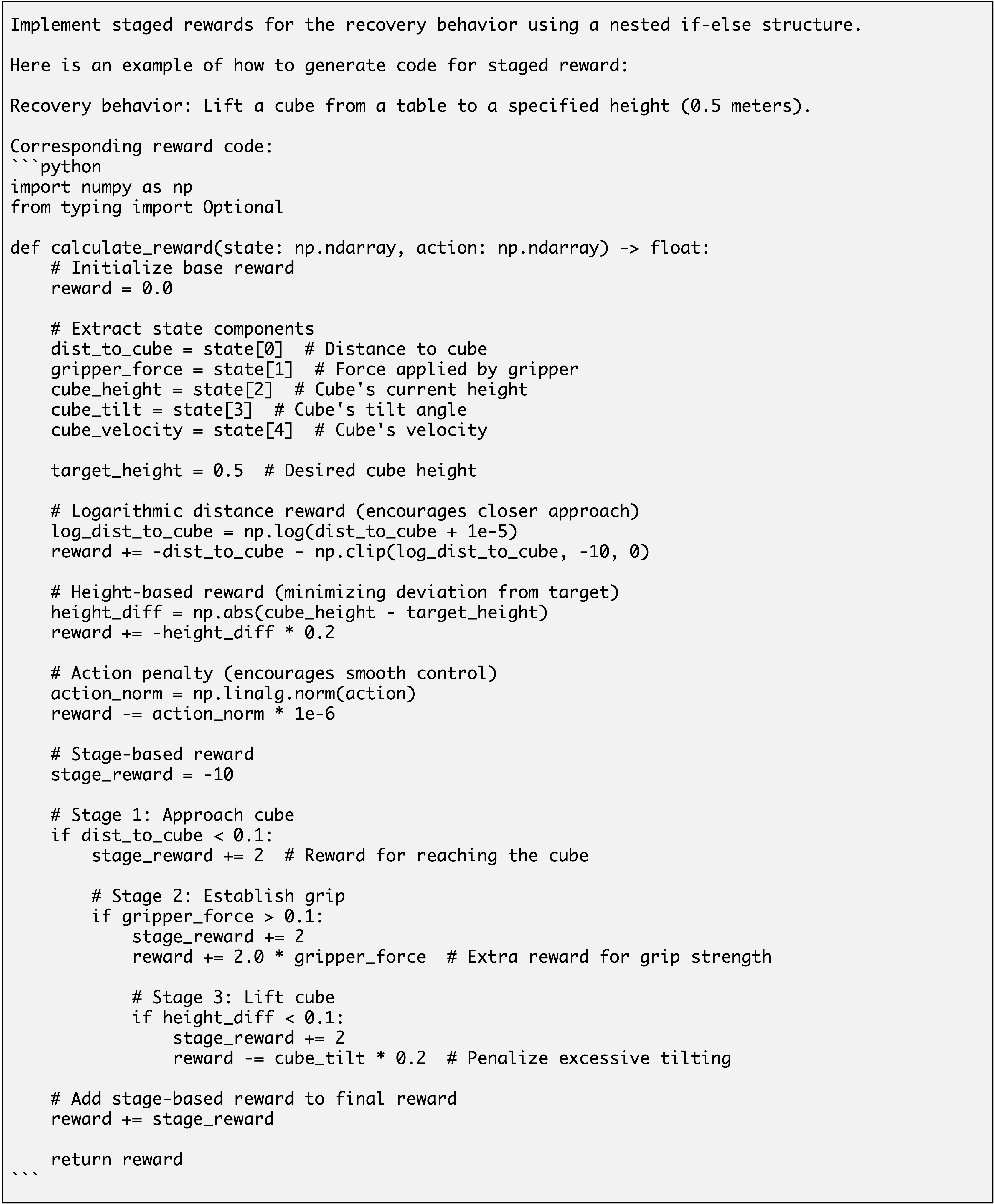}
    \caption{Few-shot example utilized in the PushChair environment.}
\label{fig_fewshot}
\end{figure*}

In our experiment, we retrain the policy in a state where the chair has already fallen to the floor. This state is an OOD state that the agent has never encountered while learning the original task, as the state lies outside the termination condition. The expected recovery behavior of the agent is to use its dual arms to grab the chair and lift it back to an upright position. The few-shot example used for \textit{Code Generation} is illustrated in Fig. \ref{fig_fewshot}.

\renewcommand{\arraystretch}{1.4}
\begin{table}[t]
    \centering
    \caption{Environmental Details}\vspace{-1em}
    \label{table_env_detail}
    \resizebox{\linewidth}{!}{
    \begin{tabular}{c c c c}
        \hline
        Environment & Termination Condition & OOD State & Scaling Coef. ($\lambda$) \\
        \hline
        Humanoid  & not ($1.0<h_{torso}<2.0$)  &  $h_{torso}=0.105$ & 0.05 \\
        PushChair  & $\theta_{chair} > \pi/5$  & $\theta_{chair} = \pi/2$ & 0.01   \\
        \hline
    \end{tabular}
    }
\end{table}

\subsection{Environment Description}
\begin{figure*}[t]
    \centering
    \includegraphics[width=0.8\linewidth]{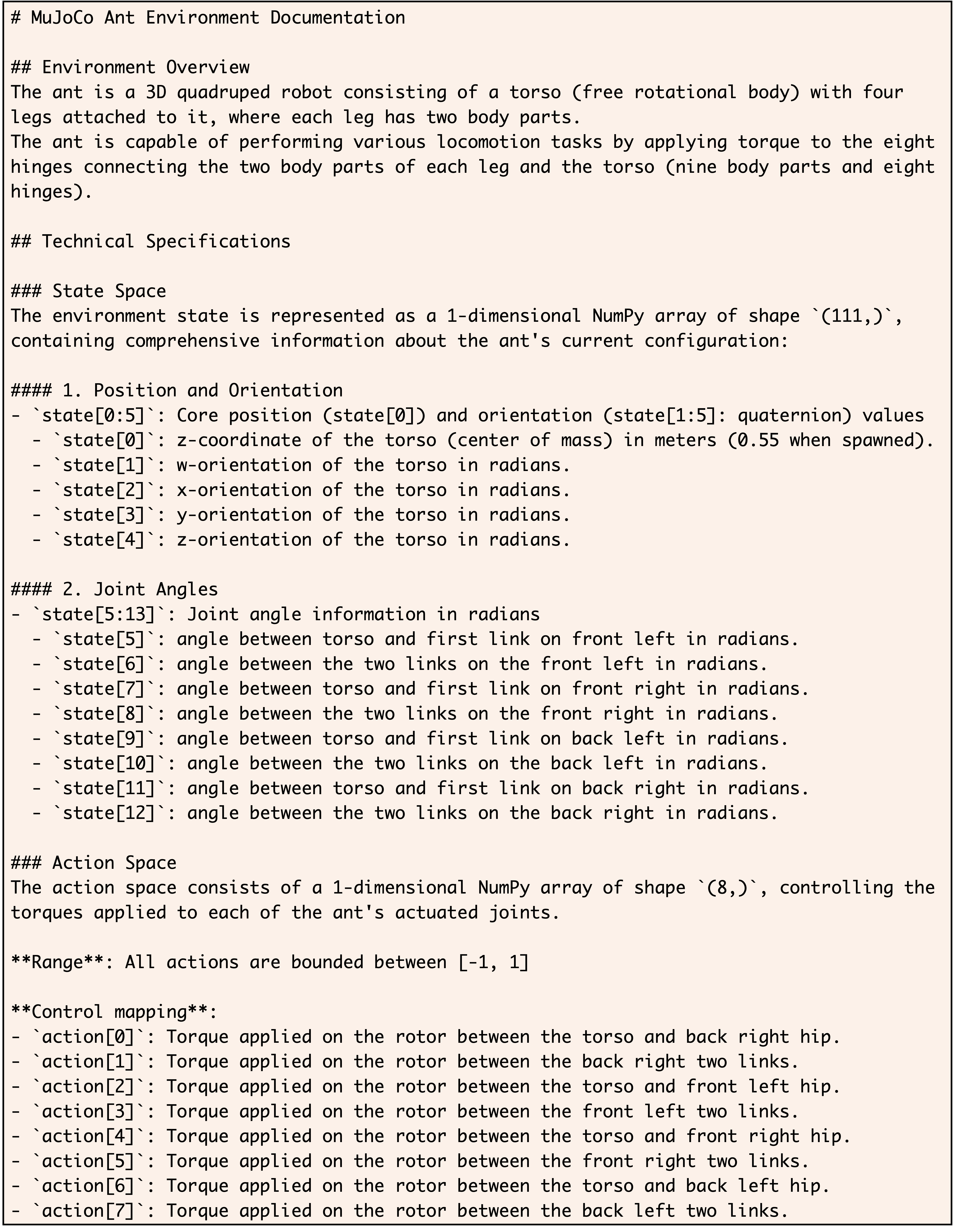}
    \caption{Environment description for the MuJoCo Ant environment.}
\label{fig_env_desc}
\end{figure*}

As outlined in the original paper, we provide environment descriptions containing state and action information to the Code Generator to generate environment-specific recovery reward codes. An example of the environment description for the MuJoCo Ant environment used in our system is shown in Fig. \ref{fig_env_desc}.

\subsection{Full Prompts}
\begin{figure*}[t]
    \centering
    \includegraphics[width=0.8\linewidth]{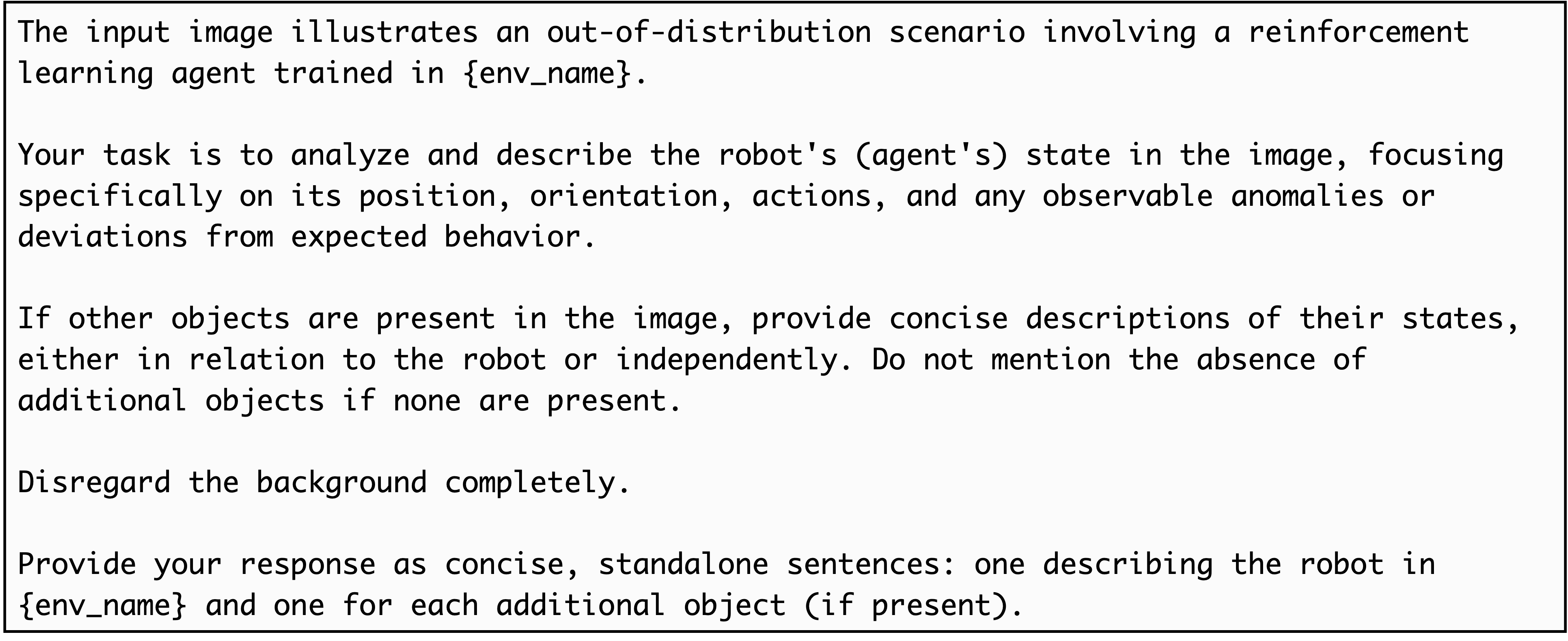}
    \caption{Prompt for the \textit{OOD Description}.}
\label{fig_ood_desc_prompt}
\end{figure*}

\begin{figure*}[t]
    \centering
    \includegraphics[width=0.8\linewidth]{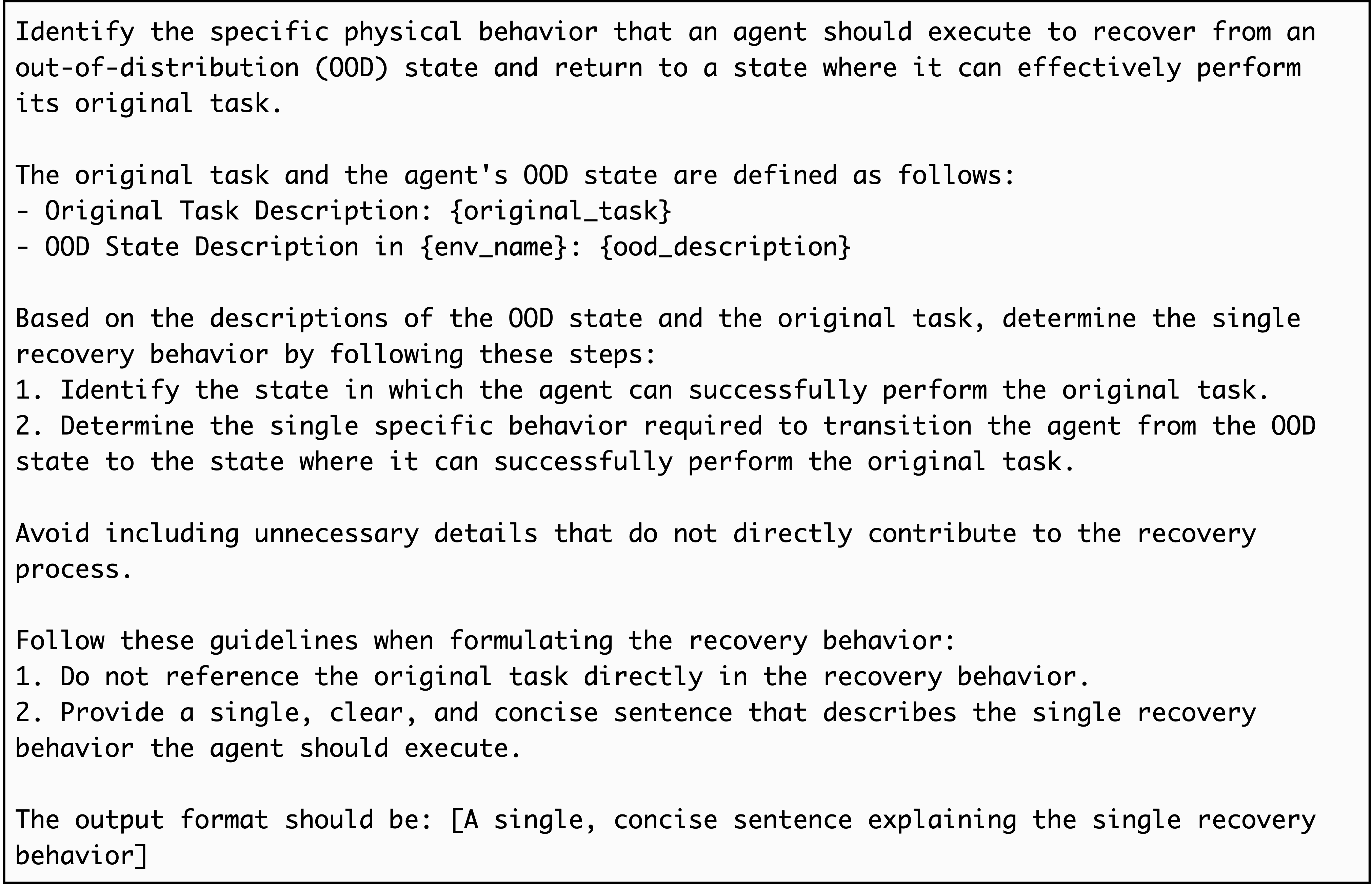}
    \caption{Prompt for the \textit{Behavior Reasoning}.}
\label{fig_beh_reason_prompt}
\end{figure*}

\begin{figure*}[t]
    \centering
    \includegraphics[width=0.8\linewidth]{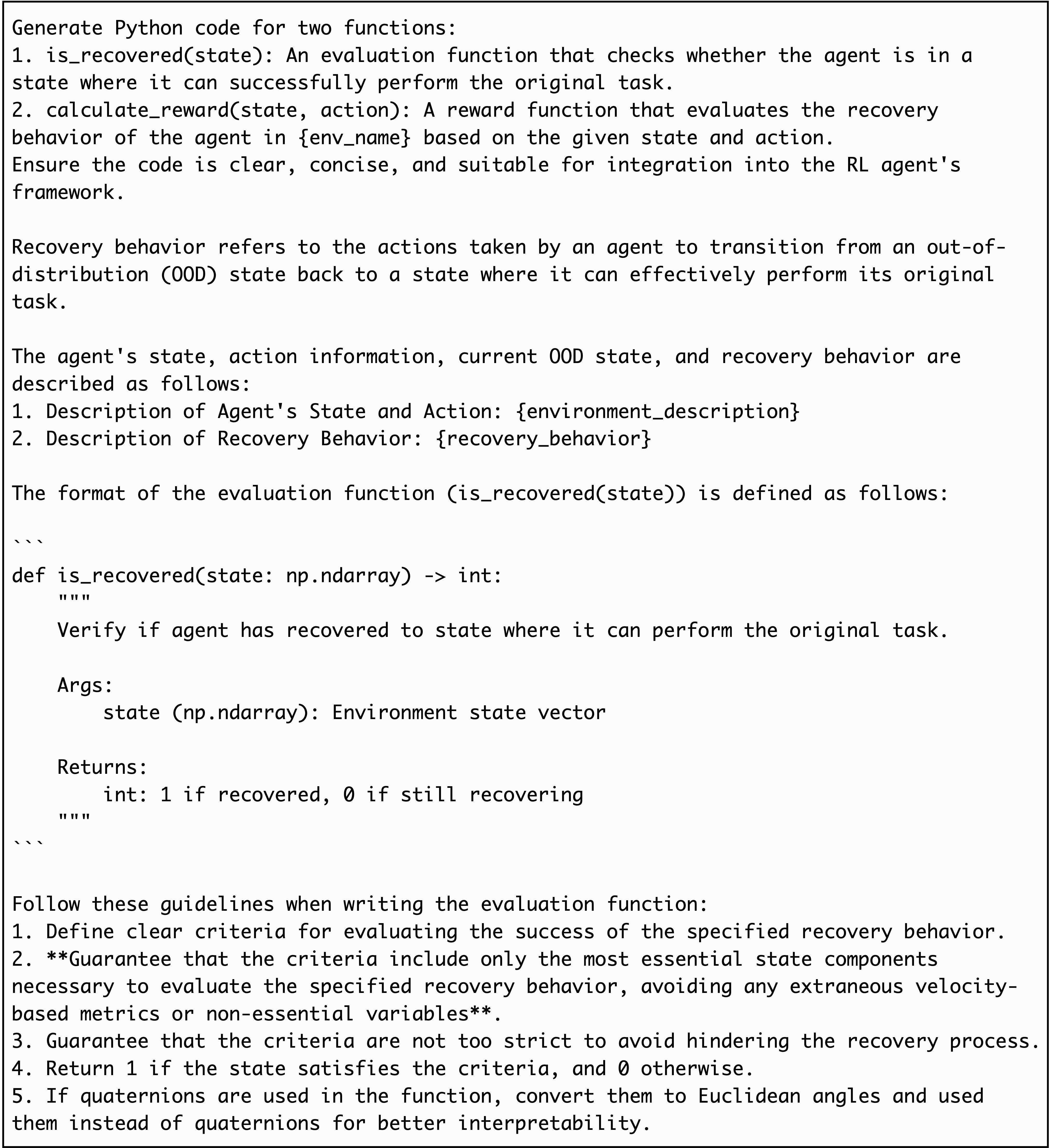}
    \caption{Prompt for the \textit{Code Generation} (1/2).}
\label{fig_code_gen_prompt1}
\end{figure*}

\begin{figure*}[t]
    \centering
    \includegraphics[width=0.8\linewidth]{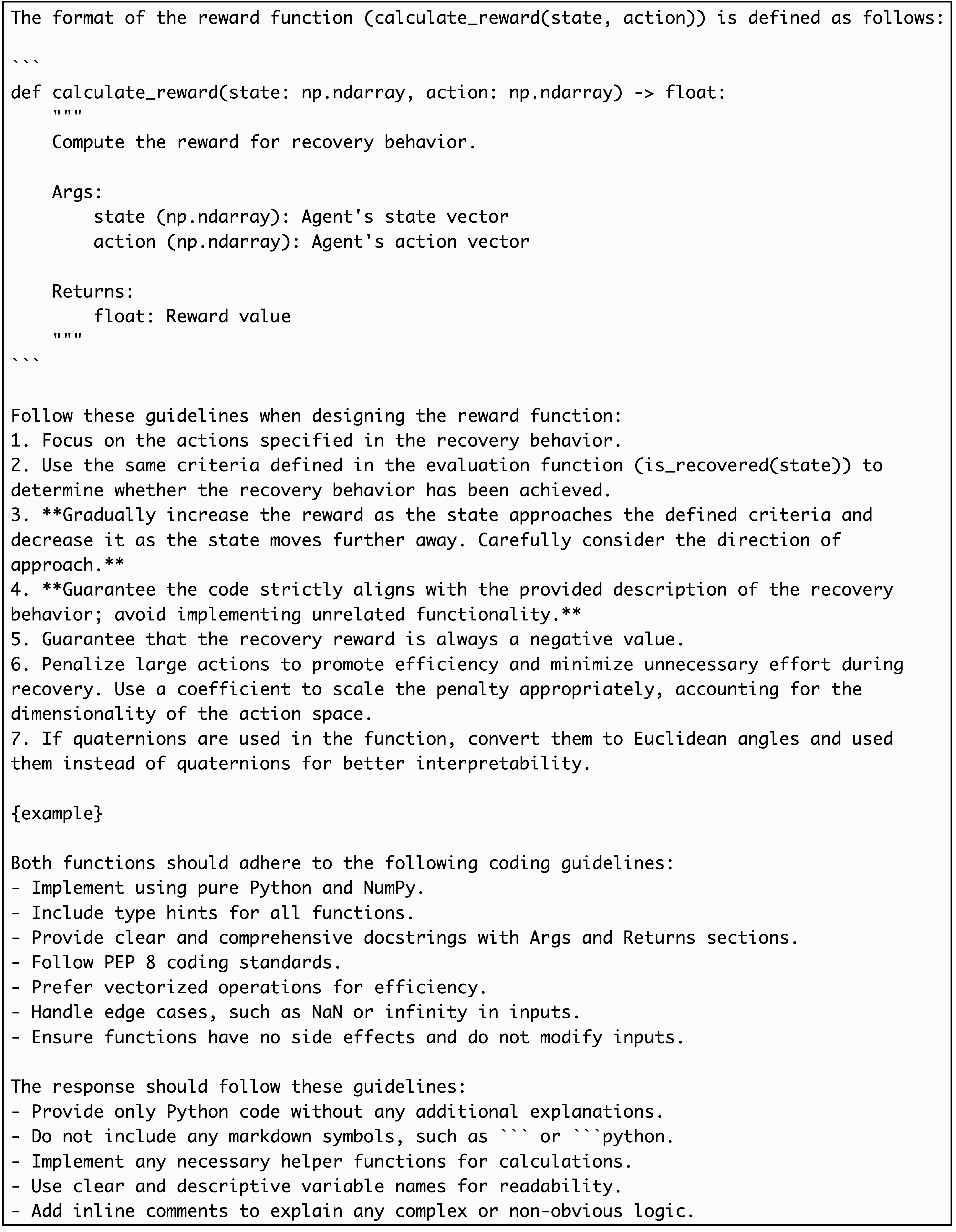}
    \caption{Prompt for the \textit{Code Generation} (2/2).}
\label{fig_code_gen_prompt2}
\end{figure*}

We include all the prompts used for our system in Fig. \ref{fig_ood_desc_prompt}--\ref{fig_code_gen_prompt2}.
\begin{itemize}
    \item \textbf{\textit{OOD Description}}: Fig. \ref{fig_ood_desc_prompt}
    \item \textbf{\textit{Behavior Reasoning}}: Fig. \ref{fig_beh_reason_prompt}
    \item \textbf{\textit{Code Generation}}: Fig. \ref{fig_code_gen_prompt1}, \ref{fig_code_gen_prompt2}
\end{itemize}

\end{document}